\documentclass{article}

   \usepackage[nonatbib, final]{neurips_2022}

\usepackage[utf8]{inputenc} 
\usepackage[T1]{fontenc}    
\usepackage{hyperref}       
\usepackage{url}            
\usepackage{booktabs}       
\usepackage{amsfonts}       
\usepackage{nicefrac}       
\usepackage{microtype}      
\usepackage{xcolor}         

\usepackage{cite}
\usepackage{graphicx}
\usepackage{subcaption}
\usepackage{amsmath}
\usepackage{amsfonts}
\usepackage[ruled,vlined]{algorithm2e}

\graphicspath{ {images/} }

\title{Reinforcement Learning-Based Air Traffic Deconfliction}

\author{%
  Denis Osipychev\\
  Boeing Research \& Technology\\
  Huntsville, AL\\
  \texttt{denis.osipychev@boeing.com} \\
  \And
  Dragos Margineantu \\
  Boeing Research \& Technology \\
  Seattle, WA\\
  \texttt{dragos.margineantu@boeing.com} \\
  \And
  Girish Chowdhary \\
  University of Illinois at Urbana-Champaign \\
  girishc@illinois.com \\
}

\begin{document}

\maketitle

\begin{abstract}
Remain Well Clear, keeping the aircraft away from hazards by the appropriate separation distance, is an essential technology for the safe operation of uncrewed aerial vehicles in congested airspace.
This work focuses on automating the horizontal separation of two aircraft and presents the obstacle avoidance problem as a 2D surrogate optimization task. By our design, the surrogate task is made more conservative to guarantee the execution of the solution in the primary domain. 

Using Reinforcement Learning (RL), we optimize the avoidance policy and model the dynamics, interactions, and decision-making. 
By recursively sampling the resulting policy and the surrogate transitions, the system translates the avoidance policy into a complete avoidance trajectory. Then, the solver publishes the trajectory as a set of waypoints for the airplane to follow using the Robot Operating System (ROS) interface.

The proposed system generates a quick and achievable avoidance trajectory that satisfies the safety requirements. 
Evaluation of our system is completed in a high-fidelity simulation and full-scale airplane demonstration.
Moreover, the paper concludes an enormous integration effort that has enabled a real-life demonstration of the RL-based system.
\end{abstract}

\let\thefootnote\relax\footnotetext{This research was developed with funding from the Defense Advanced Research Projects Agency (DARPA). The views, opinions and/or findings expressed are those of the author and should not be interpreted as representing the official views or policies of the Department of Defense or the U.S. Government.
}

\section{Introduction}
\label{sec:intro}

An automated conflict resolution is crucial for the safe operation of uncrewed aerial vehicles (UAVs) and Air Traffic Control (ATC)\cite{calandrillo2020deadly, macpherson2018world}. To guarantee safety in shared airspace, Federal Aviation Administration (FAA) specifies the requirements for UAV behavior in the following concepts \cite{faa_nas}: 
\begin{itemize}
    \item Detect and Avoid (DAA) -- detecting and avoiding hazards in the planning and execution phases of the flight;
    \item Remain Well Clear -- keeping the aircraft away from hazards by at least the appropriate separation distance.
\end{itemize}

The regulator requires all flying vehicles to maintain a minimum safe distance by providing a vertical or horizontal separation. 
The most common way is vertical separation -- assigning different flight levels. This approach may be ineffective for heavy traffic congestion, aircraft with a limited energy level, or due to extended reaction time.
The horizontal separation is more complex and requires synchronization of actions generally provided through guidance from the ATC or by the set of rules prescribing specific behavior based on the type or position of the aircraft.

In this work, we focus on horizontal separation and develop a fast-acting solution for the case that today's methods do not resolve: horizontal separation in uncontrolled airspace with no procedural separation.

\begin{figure}[h]
    \begin{subfigure}{0.45\linewidth}
        \hspace*{-10pt}
        \includegraphics[width=\linewidth]{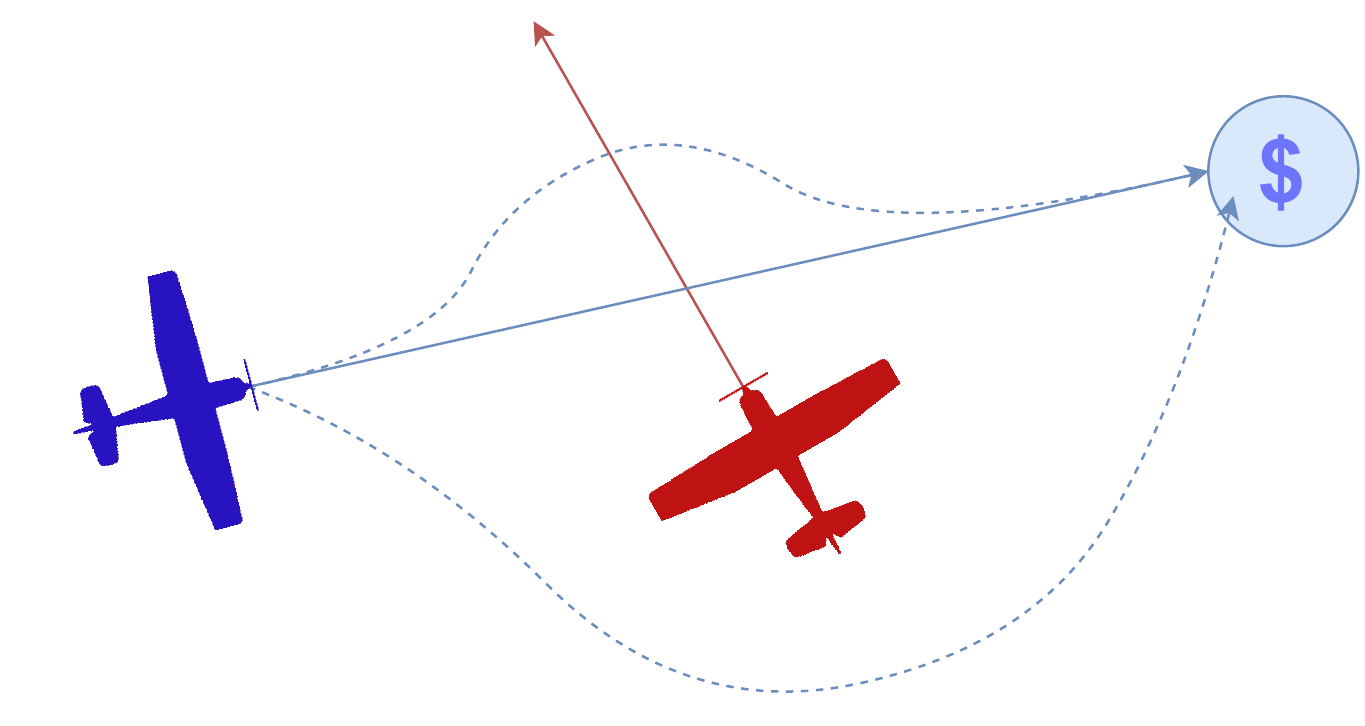}
        \caption{Horizontal separation of two aircraft formulated as a 2D car-like obstacle avoidance problem.}
        \label{fig:intro}
    \end{subfigure}
    \hspace*{10pt}
    \begin{subfigure}{0.5\linewidth}
        \hspace*{-15pt}
        \includegraphics[width=1.1\linewidth]{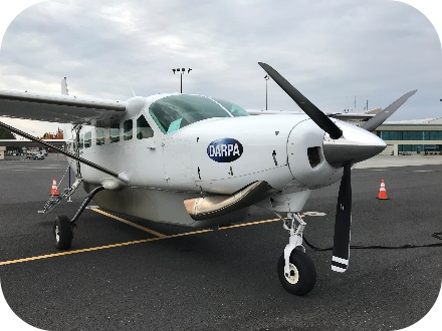}
        \caption{Single-engine turboprop Cessna 208 Grand Caravan  -- our demonstration platform.}
        \label{fig:caravan}
    \end{subfigure}
    \caption{Air traffic conflicts can be resolved by generating an avoidance trajectory that provides a safe distance between the airplanes.}
\end{figure}

This work poses the deconfliction problem as a 2D dynamic path-planning problem. The optimization is a data-driven, simulation-based sequential decision-making process using deep Reinforcement Learning (RL). This formulation allows managing the stochastic nature of the problem often overlooked by energy-based and graph-based trajectory methods \cite{kuwata2008motion, warren1989global}. 
With the amount of work in the deconfliction and collision avoidance areas for autonomous car driving, this work stands out by focusing specifically on aerial application \cite{woo2020collision, kaushik2018overtaking, zhao2019control, osipychev2017human}. Unlike automobile applications, the solution space is bounded only by the dynamics of the aircraft, which increases the search space.

As shown in Figure \ref{fig:intro}, the goal is to alternate the conflicting course of the controlled airplane (Agent), provide a safe distance to another aircraft (Intruder), and return to the original route when safe.
This work places the following strong assumptions, which may be lifted in future research.
First, the Intruder is limited to a single aircraft flying a passive trajectory -- not changing its intent and not reacting to any action of the Agent.
Second, the Agent's and Intruder's states are fully observable and assume the perfect knowledge of the limited set of parameters. This narrow set is enough to describe the Markovian state of the Markov Decision Processes (MDP) system.

This work proposes using a pre-computed avoidance policy solved for a surrogate task to improve response time and provide a run-time solution. The surrogate problem is designed to be more conservative to guarantee the feasibility of the solution in the actual task. The output of our system provides a complete avoidance trajectory as a set of waypoints for the airplane to follow. This format is preferred in the aerospace industry and is easier to integrate into the existing autopilot functionality. Compared to the end-to-end policy controlling the ailerons, rudder, and throttle, our approach would provide a visually explainable solution that can be validated before application.

The proposed system is designed to provide a quick and feasible avoidance trajectory that satisfies the safety requirements given the uncertainty in the transitions.
Evaluation of the efficiency of our Reinforcement Learning conflict avoidance system (RLCAS) is completed on the actual task in high-fidelity simulation and a full-scale airplane demonstration flying Cessna Caravan 208 shown in Figure \ref{fig:caravan}.
\section{Methodology}
\label{sec:method}
\subsection{Technical Approach}

In this work, the corrective action for the avoidance maneuver is generated by a pre-computed solution -- the Reinforcement Learning (RL) policy model. 
The RL is a sequential policy optimization method that requires continuous data-rich interaction with the environment \cite{sutton2018reinforcement}. 
We trained the policy on a surrogate rather than the original task to provide the required number of interactions. 
The surrogate environment developed by Boeing is a lightweight Python environment integrated with OpenAI GYM framework \cite{brockman2016openai}.

\begin{figure}[h]
   \centering
   \includegraphics[width=0.8\linewidth]{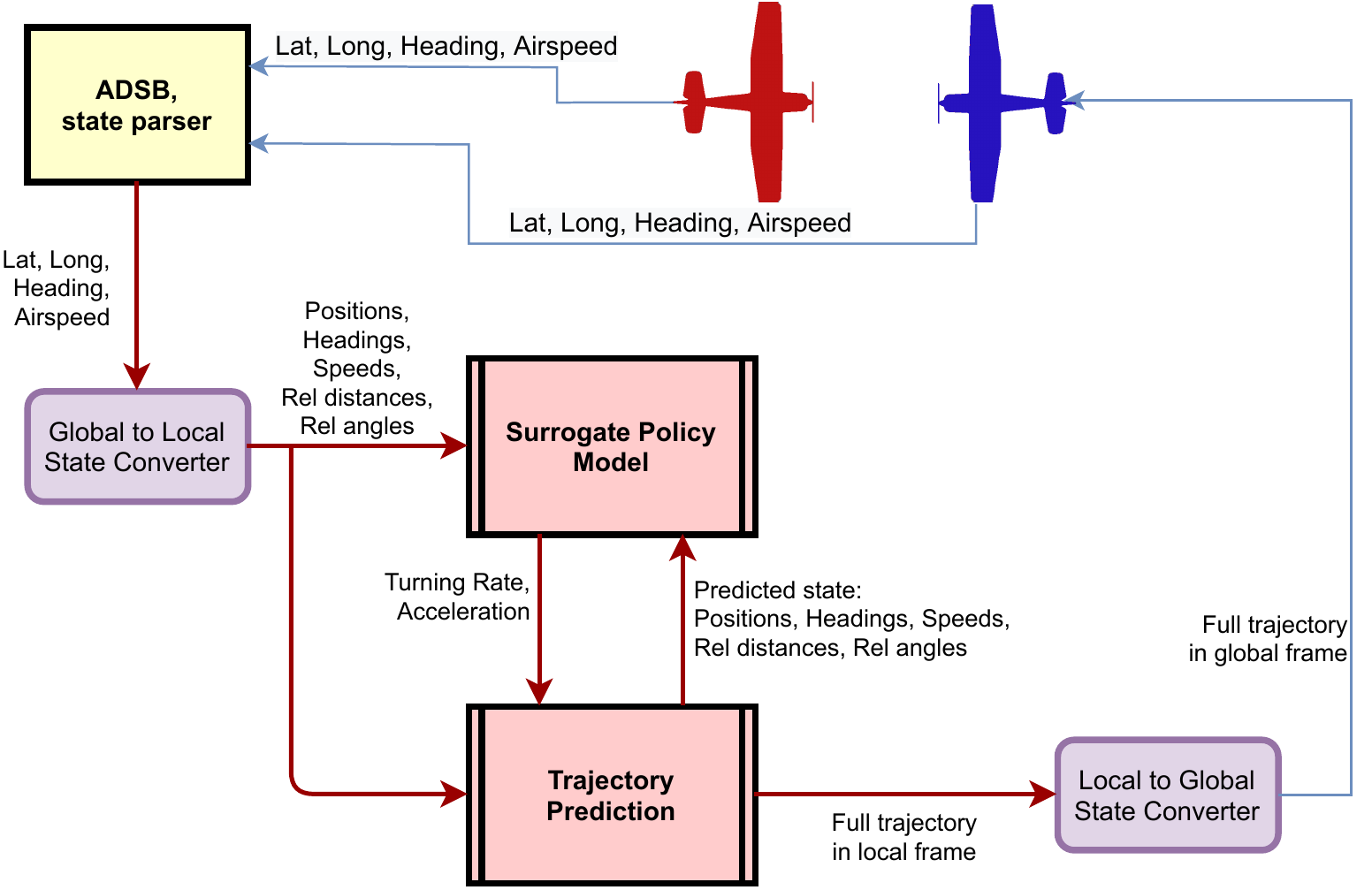}
    \caption{System diagram of the proposed model-based conflict resolver.}
    \label{fig:diagram}
\end{figure}

The learned model minimizes the risk by providing continuous control commands in the surrogate environment. These commands are translated into a geometric trajectory using the surrogate environment.
The surrogate policy $\hat{\pi}$ is a data-driven end-to-end RL policy for the surrogate task that shares similar transition dynamics $\hat{T}$ and rules of engagement $\hat{R}$. The surrogate policy is applied to the original task assuming the similarity of the transitions and conditions. The surrogate is a more conservative approximation of the actual task that guarantees the solution will be feasible in the real system.

\begin{align}
    \pi_{\text{True}} &= \pi(R_{\text{True}}, T_{\text{True}}) \\
    \hat{\pi} &= \pi^*(\hat{R}, \hat{T}) \\
     \pi_{\text{True}} &\approx \hat{\pi} \text{, if } 
     [ R_{\text{True}}, T_{\text{True}} ] \in [\hat{R} + \epsilon, \hat{T} + \epsilon ]
\end{align} 

\subsection{Surrogate Environment}

\begin{table}[ht]
    \centering
    \begin{tabular}{|| c | c | c | c ||} 
    \hline
    Param & Surrogate & C208 (True) & Units \\
    \hline
    X & -10000 .. 10000 & .. & m \\
    Y & -10000 .. 10000 & .. & m \\
    Heading & -180 .. 180  & -180 .. 180 & deg \\
    Airspeed & 50 .. 100 & 31 .. 100  & m/s \\
    \hline
    Long accel. & -0.5 .. 0.5 &  -0.8 .. 0.5 & m/s$^2$ \\
    Yaw rate & -3 .. 3 & -5 .. 5  & deg/s \\
    \hline
    \end{tabular}
    \caption{Surrogate vs. true dynamic parameters and ranges.}
    \label{tbl:sur_true_compar}
\end{table}

Our surrogate environment is a simplified 2D obstacle avoidance problem that mimics the actual task (air traffic conflict resolution). 
The aircraft control differs from 2D car-like dynamics and does not have conventional brakes, acceleration, or steering. The effectiveness of the controls depends on the aircraft's altitude, airspeed, etc. The parameters selected for turning rates and longitudinal accelerations are conservative and guaranteed across the full flying envelope of the picked aircraft. Selected longitudinal accelerations allow slight deviations in the aircraft's altitude within the desired flight level.

The turning rates have been selected to guarantee the aircraft will be capable of performing the required turning radius. Table \ref{tbl:sur_true_compar} outlines the difference in the main dynamic parameters of the simulated vehicle. A high-fidelity simulation of this specific airframe has been used to evaluate the surrogate policy performance. Detailed vehicle response to the commanded actions can be found in the Appendix in Figures \ref{fig:accel_data}, \ref{fig:decel_data}. All the observation and control values were normalized to [-1..1] for the convenience of the RL training.

The environment simulates the movements of two aircraft in a 20x20 kilometers area. The controlled Agent has to go around the Intruder, provide minimal horizontal separation, and merge back to the next safe waypoint from the original route before the simulation ends. 
Detailed parameters of the training simulation are provided in the Appendix in Table \ref{tbl:sim_param}.
The surrogate provides sparse reward feedback for the interactions shown in greater detail in the Appendix in Table \ref{tbl:sim_reward}.

The simulation update step follows the OpenAI GYM convention while triggering the following behavior: convert the commands into input values, set the inputs to the vehicles, step vehicles' dynamics, then evaluate the conflicts. The detailed algorithm is shown in the Appendix Algorithm \ref{alg:updates}.

A simplistic dynamic model represents all the vehicles in the surrogate:
\begin{itemize}
\item Dubin's vehicle model for turn dynamics
\item Mass-less kinematic model 
\end {itemize}

To facilitate the training, we created a simplified PID-based waypoint controller for the Intruder vehicle.
In the beginning, we do not anticipate the Intruder changing its direction or speed. 
The waypoint controller primarily helps facilitate the training and generate conflicts in training scenarios.
However, changing the Intruder's intent is planned for future work.

\subsection{RL Training}

The training wrapper allows the override of certain environment methods and parameters and introduces certain important changes to the environment, making policy training much more efficient. This wrapper includes:
\begin{enumerate}
    \item scoring,
    \item initialization - vehicles have a high chance of conflicting trajectories,
    \item repacked and normalized observations.
\end{enumerate}

To ensure the Agent generalizes the problem, we rewrap the observations and focus only on relative positions rather than absolute ones. In addition, we normalize the values to [-1 .. 1].
Repacked observations consumed by the Agent:
\begin{center}
\begin{tabular}{|| c | c ||} 
\hline
heading & intruder heading\\
airspeed & intruder airspeed\\
distance to goal & distance to intruder\\
tracking angle to goal & tracking angle to intruder\\
\hline
\end{tabular}
\end{center}

\begin{figure}[h]
    \begin{subfigure}{0.5\linewidth}
        \hspace*{-20pt}
        \includegraphics[width=\linewidth]{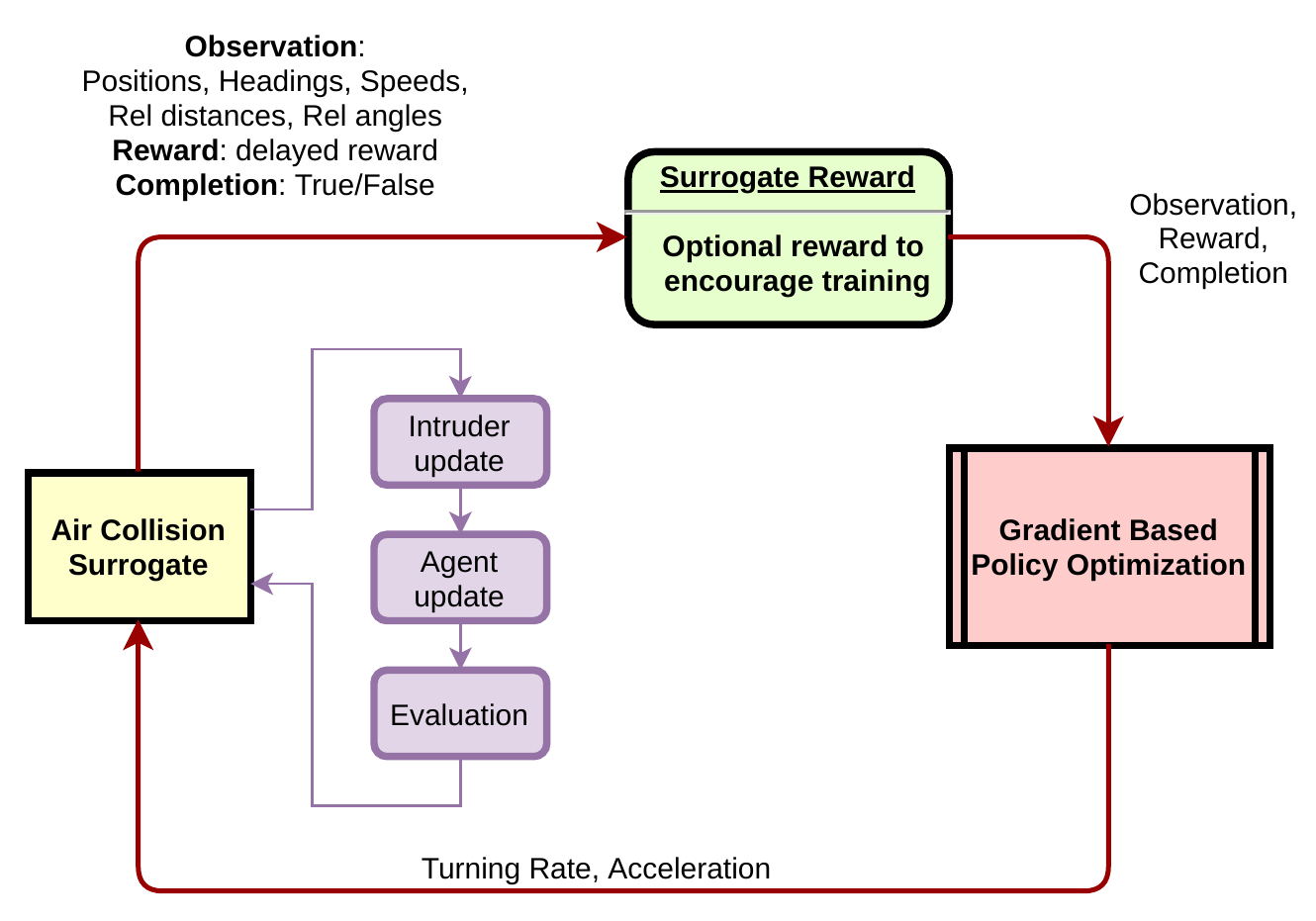}
        \caption{Reinforcement Learning (RL) framework used to optimize the surrogate policy by iteratively sampling and improving the actions.}
        \label{fig:rl_framework}
    \end{subfigure}
    \hspace*{10pt}
    \begin{subfigure}{0.45\linewidth}
        \hspace*{-5pt}
        \includegraphics[width=\linewidth]{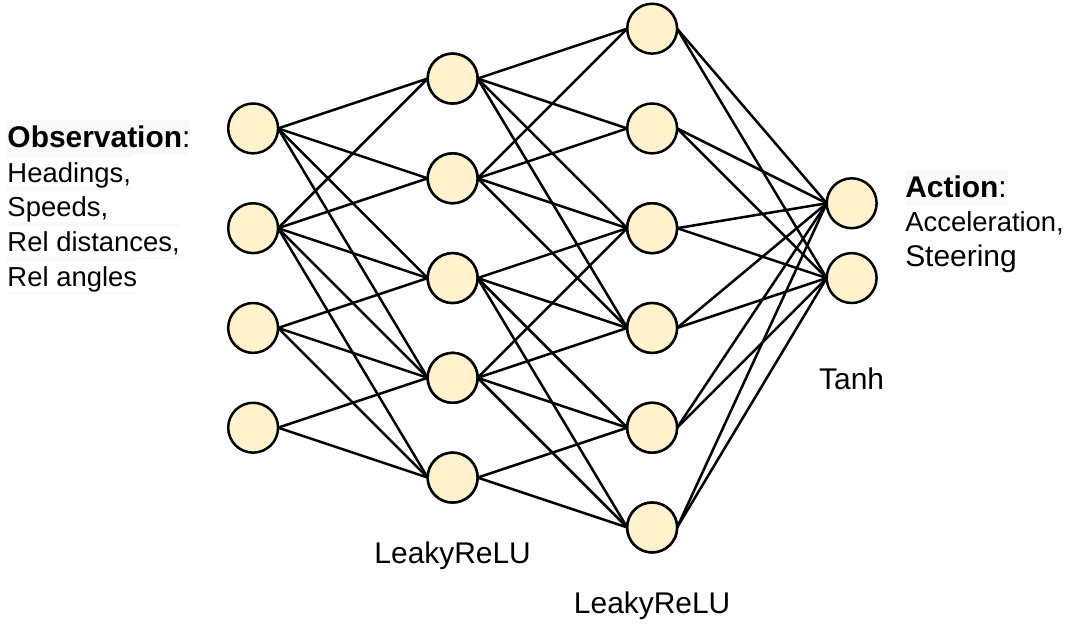}
        \caption{Neural network function approximation used for the policy model consists of 2 hidden layers, 256 neurons each.}
        \label{fig:policy_model}
    \end{subfigure}
    \caption{Reinforcement Learning policy.}
\end{figure}

To better cover the configuration space, we modified the initialization routine. The initial positions are sampled from an imaginary circle, with both vehicles moving toward the circle's center. This guarantees that their paths intersect and the vehicles are synchronized. As a result, the Agent and Intruder have a higher chance of conflicting trajectories.

\subsection{RL Policy Agent}

The RL policy agent learns the task by interacting with the simulation and iteratively updating the parameters of the policy model using Stochastic Gradient Descent (SGD) optimization. We approximate the policy with a multi-layered perceptron shown in Figure \ref{fig:policy_model}.

To solve the optimization problem as Markov Decision Problem (MDP) system, we refactor it into Markovian states $s$, transitions  $T(s' | s, a)$, and transition reward $R(s' | s, a)$. 
The system's state (including both Agent and Intruder) is fully observable, assumes the perfect knowledge, and is enough to describe the Markovian state of the MDP system.
The state of the agent described as
$$ s = \{ v_a, \psi_a, v_i, \psi_i, \beta_i, d_i, \beta_g, d_g \}$$

where 
    $v_a$ - Agent speed,
    $\psi_a$ - Agent heading,
    $v_i$ - Intruder speed,
    $\psi_i$ - Intruder heading,
    $\beta_i$ - angle to Intruder,
    $d_i$ - distance to Intruder,
    $\beta_g$ - angle to goal,
    $d_g$ - distance to the goal. 

The optimization is set to find the optimal policy $\pi^*(s)$ as a set of state-action mappings that maximizes the expected reward $V(s)$ \cite{sutton2018reinforcement}.
\begin{align} 
    \pi(s) &= P(a | s) \\
    \pi^*(s) &= \arg\max_{\pi} V^\pi (s) \\
    &= \arg\max_a \left( R(s,a) + \gamma T(s'|s,a) V(s') \right)
\end{align} 

Value of the state is the expected future reward accumulated over the trajectory and defined by Bellman function as:
\begin{align}
    V(s) &=  \mathbb{E} [R | s, \pi] \\
    &= \sum_{s'} T(s'|s,a) \left( R(s'|s,a) + \gamma ( V(s')) \right) \\
    &=  R(s'|s,a) + \gamma \sum_{s'} T(s'|s,a) V^{\pi}(s') \\
    V^{*}(s) &= \max_{a} \left( R(s,a) + \gamma \sum_{s'} T(s'|s,a) V^{*}(s') \right)
\end{align}

The RL policy model is based on the Actor-Critic architecture that helps to improve the stability of the training \cite{sutton2018reinforcement}. The SGD-based update for Actor $\theta$ and Critic $w$ networks:
\begin{align}
    \delta &=  R_{t+1} +\gamma \hat V(s_{t+1},w) - \hat V(s_t,w) \\
    w &\leftarrow w + \alpha \delta \nabla \hat V (s, w) \\
    \theta &\leftarrow \theta + \alpha \delta \nabla \ln \pi (a|s, \theta)
\end{align} 

The core functionality of the RL agent incorporates the Stable Baselines library, a very reputable fork of OpenAI Baselines \cite{hill2018stable}. For the exploration policy and update steps, this work used the Proximal Policy Optimization (PPO) algorithm that becomes state of the art in continuous-action agents \cite{schulman2017proximal}.
\section{Integration}
\label{sec:integr}
\subsection{Training Results}
To make the RLCAS system quick and efficient in run-time, we pre-compute all the solutions beforehand and store them as a neural network model. To create this library of solutions, the neural net goes through thorough training, which is the most computationally expensive process. The training is performed in the lightweight Python surrogate environment that simplifies the dynamics of the vehicles to a 2D geometric problem. Our best policy is the result of running the simulation for $10^8$ steps (around $10^6$ episodes or three years of sim time).

\begin{figure}[h]
    \begin{subfigure}{0.5\linewidth}
        \includegraphics[width=\linewidth]{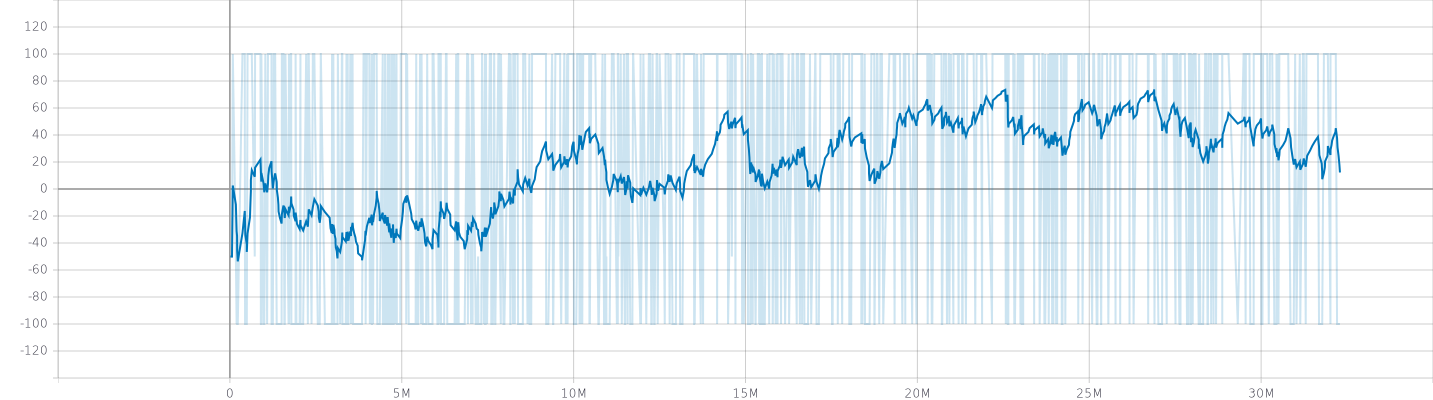}
        \caption{Average episode reward}
        \label{fig:result01}
    \end{subfigure}
    \begin{subfigure}{0.5\linewidth}
        \includegraphics[width=\linewidth]{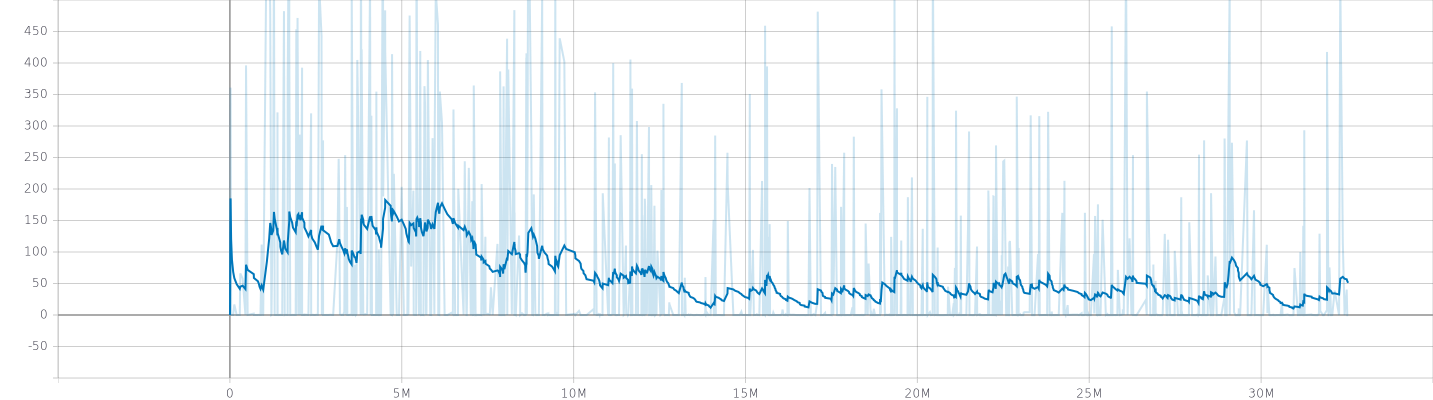}
        \caption{Training Loss}
        \label{fig:result02}
    \end{subfigure}
    \\
    \begin{subfigure}{0.5\linewidth}
        \includegraphics[width=\linewidth]{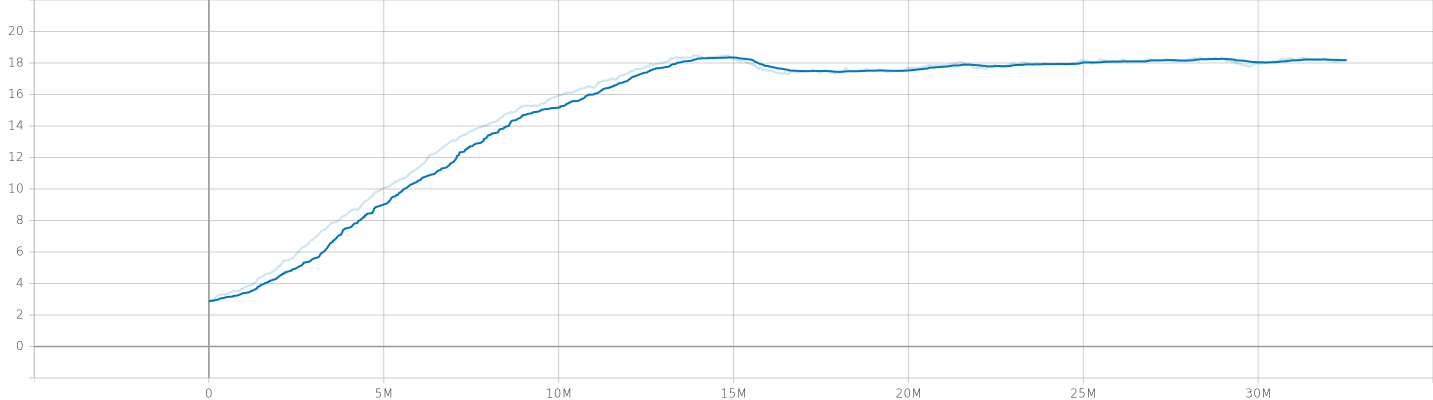}
        \caption{Entropy Loss}
        \label{fig:result03}
    \end{subfigure}
    \begin{subfigure}{0.5\linewidth}
        \includegraphics[width=\linewidth]{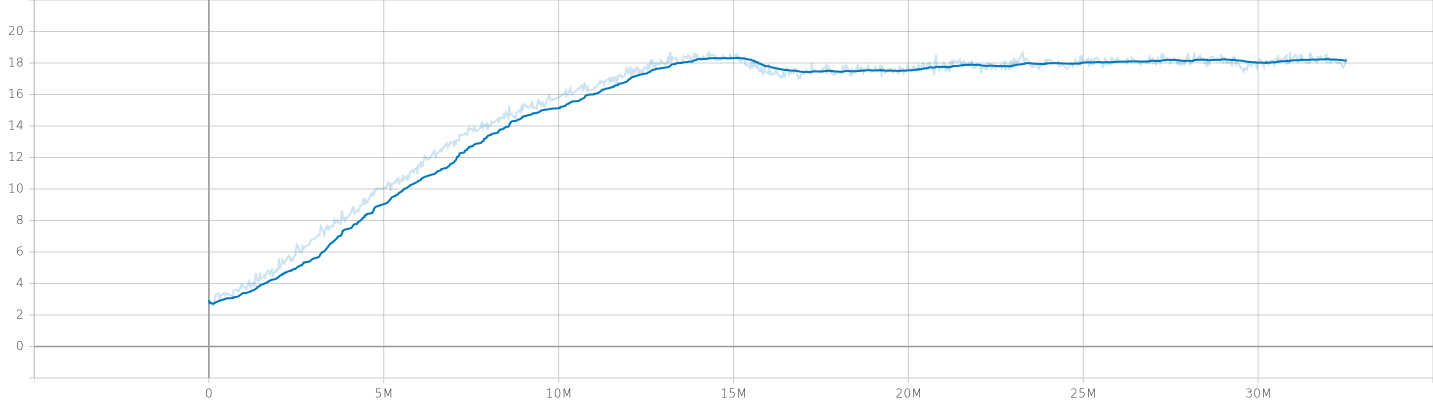}
        \caption{Neg-log Probability}
        \label{fig:result04}
    \end{subfigure}
    
    \caption{RL agent training results.}
\end{figure}

The policy converges to an average score of $+75.0$ on the surrogate task. 
The resulting policy is stored in a file that includes all the parameters of the model.
This file will be loaded and run by our ROS-Agent Runner -- an integration tool that incorporates ROS interface to communicate with the simulations and hardware.

\subsection{Evaluation and Demonstrator}

Adopting the surrogate policy to the original task requires thorough investigation. 
Due to a mismatch in the vehicle dynamics, directly wiring the policy output to the vehicle controller is unreliable, causes oscillations, and contributes to system instability.
This effect is amplified by the RL policy itself since the successful policy does not guarantee smooth and consistent actions \cite{mysore2020regularizing, cheng2019control}. 
In addition, the difference in the dynamics and the internal controller contribute to an accumulated error in the resulting trajectory. 

This deficiency is compensated by feeding the geometric trajectory to the existing vehicle controller. The trajectory represents the set of waypoints associated with an estimated arrival time. This allows the vehicle controller to compensate for the potential discrepancy in the transitions.

We evaluate the resulting policy using different demonstration platforms to prove that the policy can be safely adopted.
The work proposes the following demonstrators:
\begin{itemize}
    \item Gazebo simulation with low-fidelity dynamics,
    \item Boeing proprietary high-fidelity simulation,
    \item Real airplane demonstrator.
\end{itemize}

\subsection{ROS Integration}

\begin{figure}[h]
   \centering
   \includegraphics[width=0.8\linewidth]{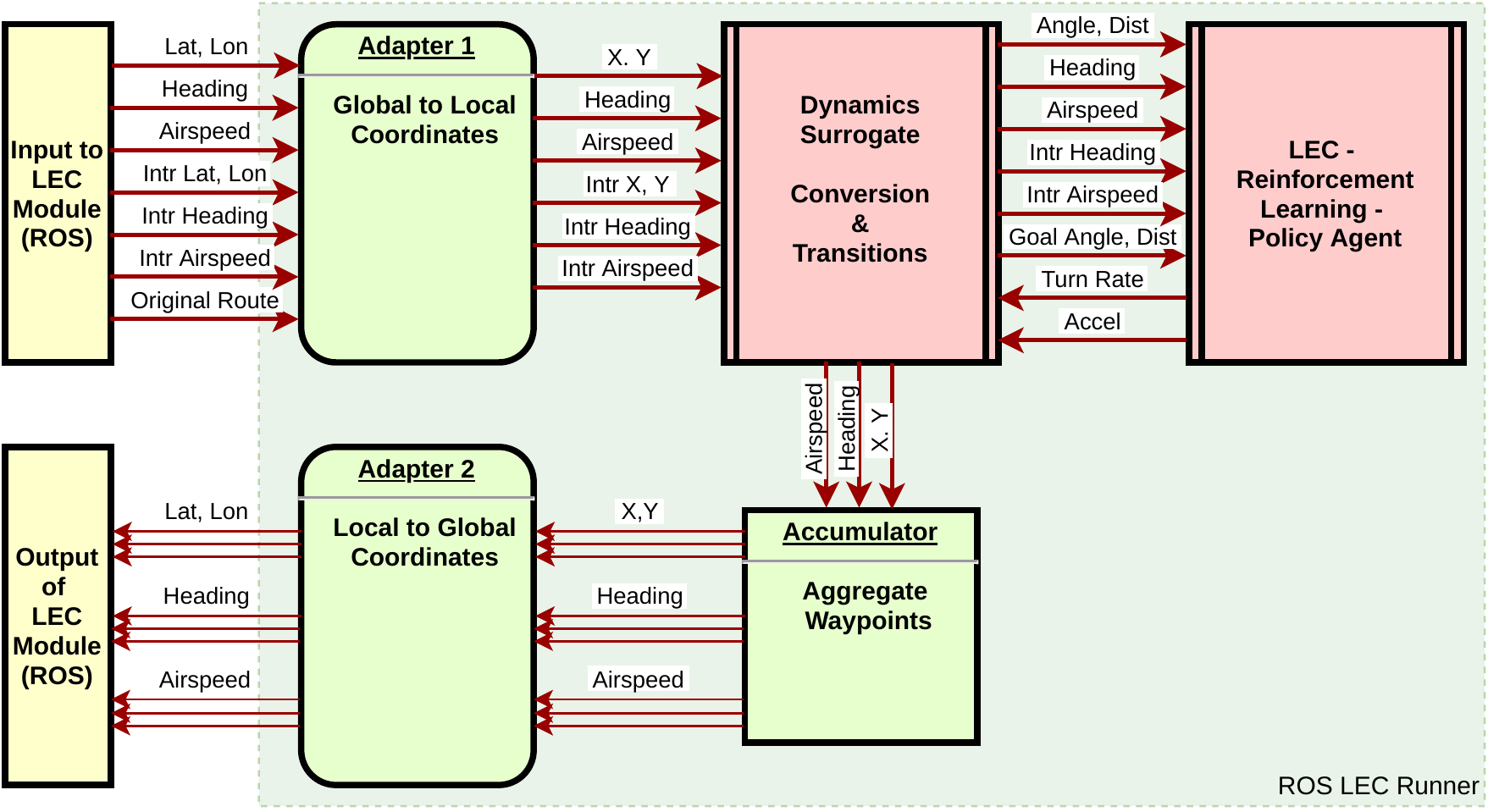}
    \caption{System diagram of the ROS-LEC (ROS-Agent) runner -- the integration of the learning-enabled component (LEC) to the physical demonstrator using ROS interface.}
    \label{fig:integration}
\end{figure}

CAS system integration is done using Robot Operating System (ROS) interface \cite{quigley2009ros}. This allows unifying the interfaces to the high-fidelity simulation and to the physical demonstrator.
ROS-Agent (ROS-LEC) runner, demonstrated in Fig. \ref{fig:integration}, aggregates data from different domains and provides important utilities to the system. Its job is designed as follows: 
\begin{itemize}
\item receive and accumulate ROS messages regarding the own-ship state,
Intruder's state, traffic alerts, GPS-SRS transformation data,
\item translate ADSB and GPS-Novatel positioning data to local coordinate frame,
\item extract the goal location from the original route,
\item re-wrap the observations into the Agent-specific input format,
\item iteratively run the Agent to get the corrective actions,
\item iteratively run the surrogate environment to receive the transitions,
\item form a corrective trajectory and check if the trajectory is good,
\item translate the trajectory from local coordinate frame to global lat-long waypoints,
\item publish the trajectory as ROS message.
\end{itemize}

On an external request, the Runner generates a single avoidance trajectory and publishes it as a ROS message. The trajectory consists of 20 waypoints in total. The last waypoint is taken from the original route, and 19 waypoints are generated by the policy. This allows linking the waypoints by a unique index and preserving the indices of the original route.

These waypoints are spaced 20 seconds apart, which provides a 400-second planning horizon. 
Because of the large time step between the waypoints, the policy and the surrogate have to be evaluated 20 times to make a single waypoint. The total response time of the system is below 60 msec for a complete trajectory.

This architecture allows closed-loop corrections with an external run-time assurance monitor. The monitor keeps track of the accumulated transition error and either request an updated avoidance plan or denies the operation switching to a backup mode. When needed to re-plan, the Runner can be requested again.
\section{Results}
\label{results}

We break the evaluation results into several parts to evaluate the complete system's performance. The first part is a statistical evaluation where we run a batch simulation to compute the total number of successful deconfliction events. The second part is the trajectory performance analysis to evaluate the numerical error introduced by the controller following the surrogate policy.

\subsection{Statistical Results}

\begin{figure}[h]
    \centering
    \includegraphics[width=0.9\linewidth]{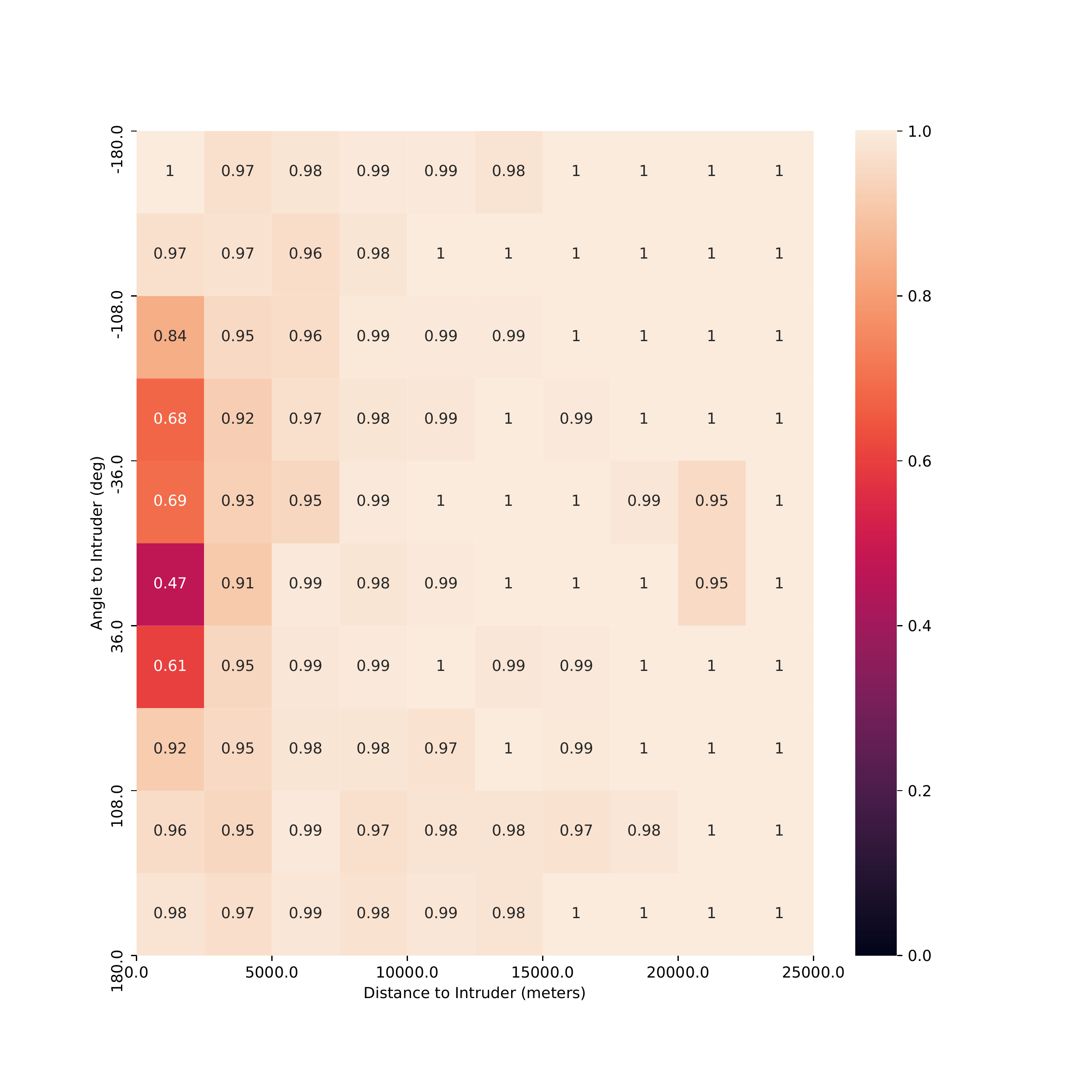}
    \caption{Statistical results demonstrate the impact of initial conditions on the system's performance. In most failures, the relative angle and distance to the Intruder presume that the conflict was inevitable.}
    \label{fig:ic_plot}
\end{figure}

To evaluate the performance of the avoidance policy itself, we have performed a batch simulation of over $10^4$ random conditions. Our RLCAS demonstrated 98\% performance in successfully avoiding the traffic conflicts with the Intruder and returning to the original route. In failed cases, 2\% of the total number of simulations, the Agent could not get back on track while maintaining a safe distance. Fig. \ref{fig:ic_plot} demonstrates that the Agent started too close to the Intruder in the cases that have not been resolved. With the relatively high number of failures, the system is not designed to work as a stand-alone system and requires a safety layer constructed on top.

\subsection{Trajectory Evaluation}

\begin{figure}[h]
    \begin{subfigure}{0.45\linewidth}
        \hspace*{-5pt}
        \includegraphics[width=1.2\linewidth]{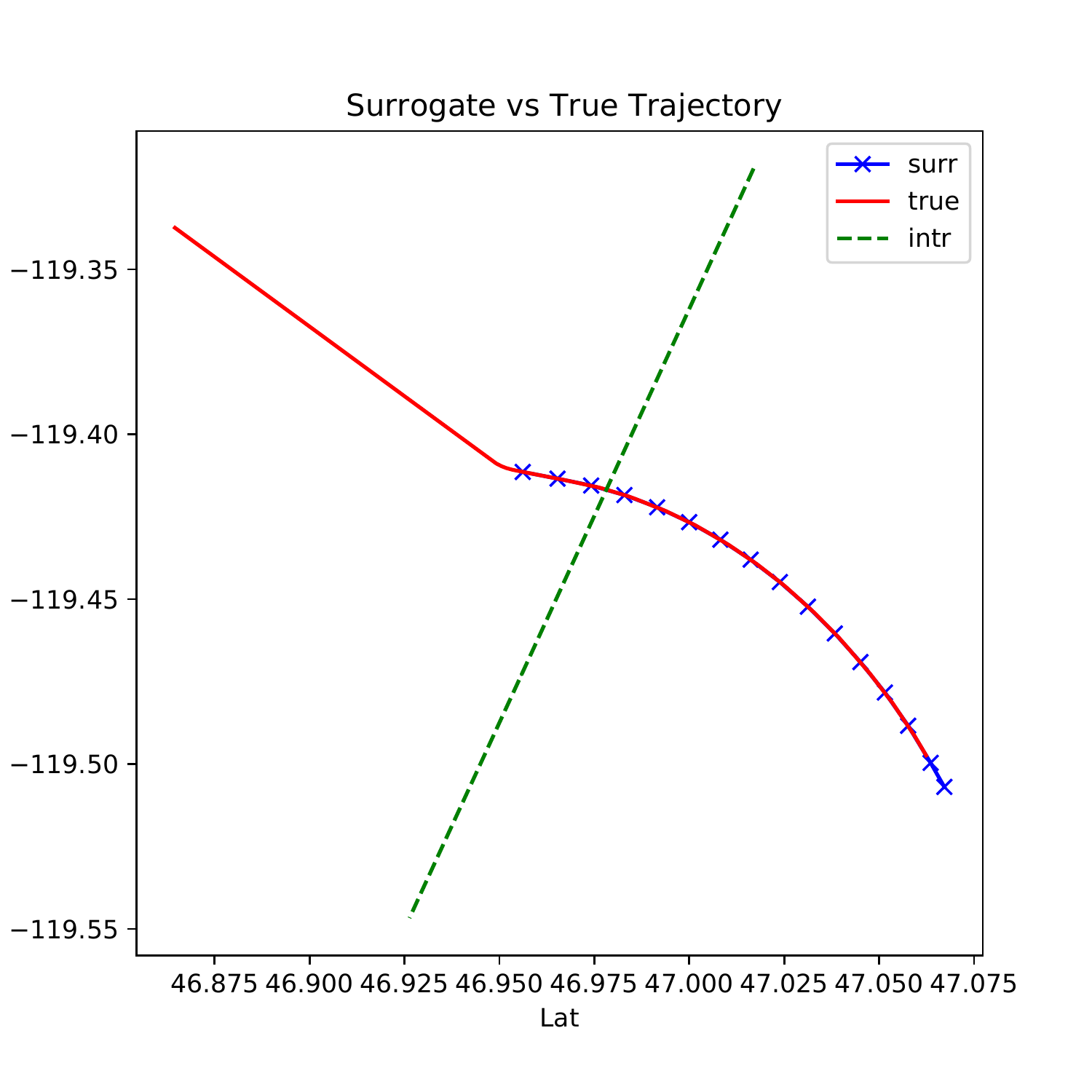}
        \caption{Intruder flying a conflicting cross course.}
        \label{fig:traj_cross}
    \end{subfigure}
    \hspace*{10pt}
    \begin{subfigure}{0.45\linewidth}
        \hspace*{-5pt}
        \includegraphics[width=1.2\linewidth]{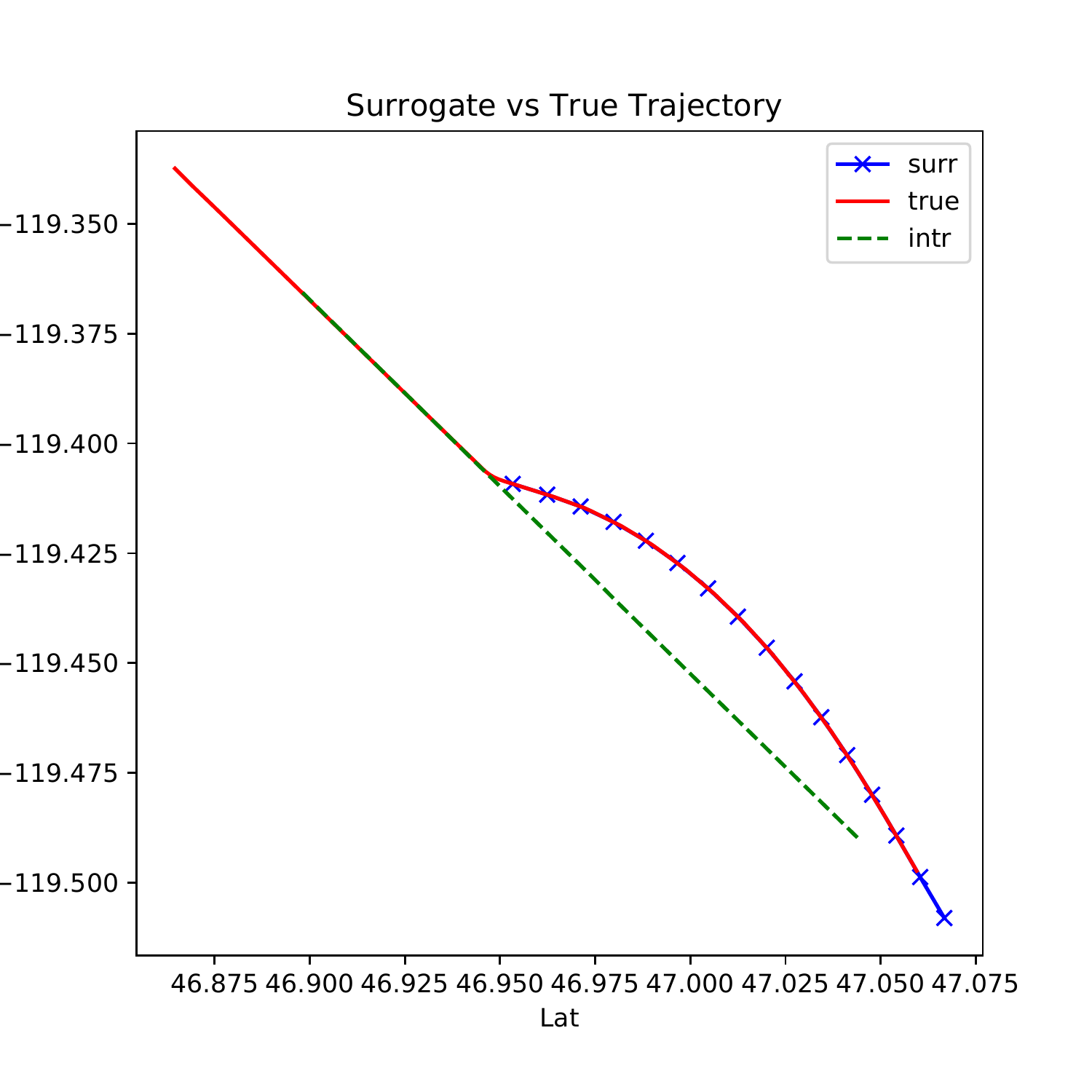}
        \caption{Intruder flying a conflicting head-on course.}
        \label{fig:traj_headon}
    \end{subfigure}
    \\
    \begin{subfigure}{0.45\linewidth}
        \hspace*{-12pt}
        \includegraphics[width=1.3\linewidth]{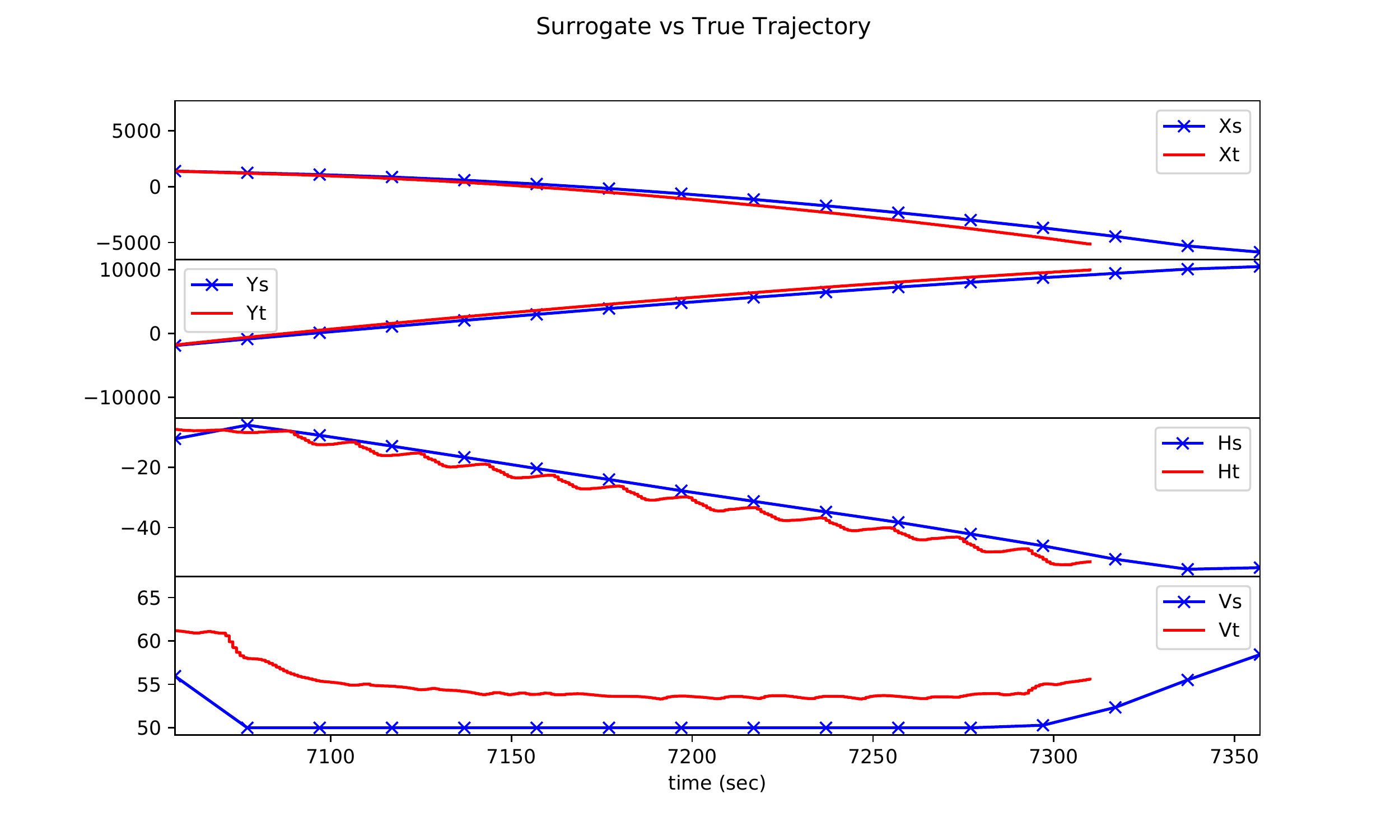}
        \caption{Accumulation of the error in the trajectory. }
        \label{fig:error_1}
    \end{subfigure}
    \hspace*{10pt}
    \begin{subfigure}{0.45\linewidth}
        \hspace*{-13pt}
        \includegraphics[width=1.3\linewidth]{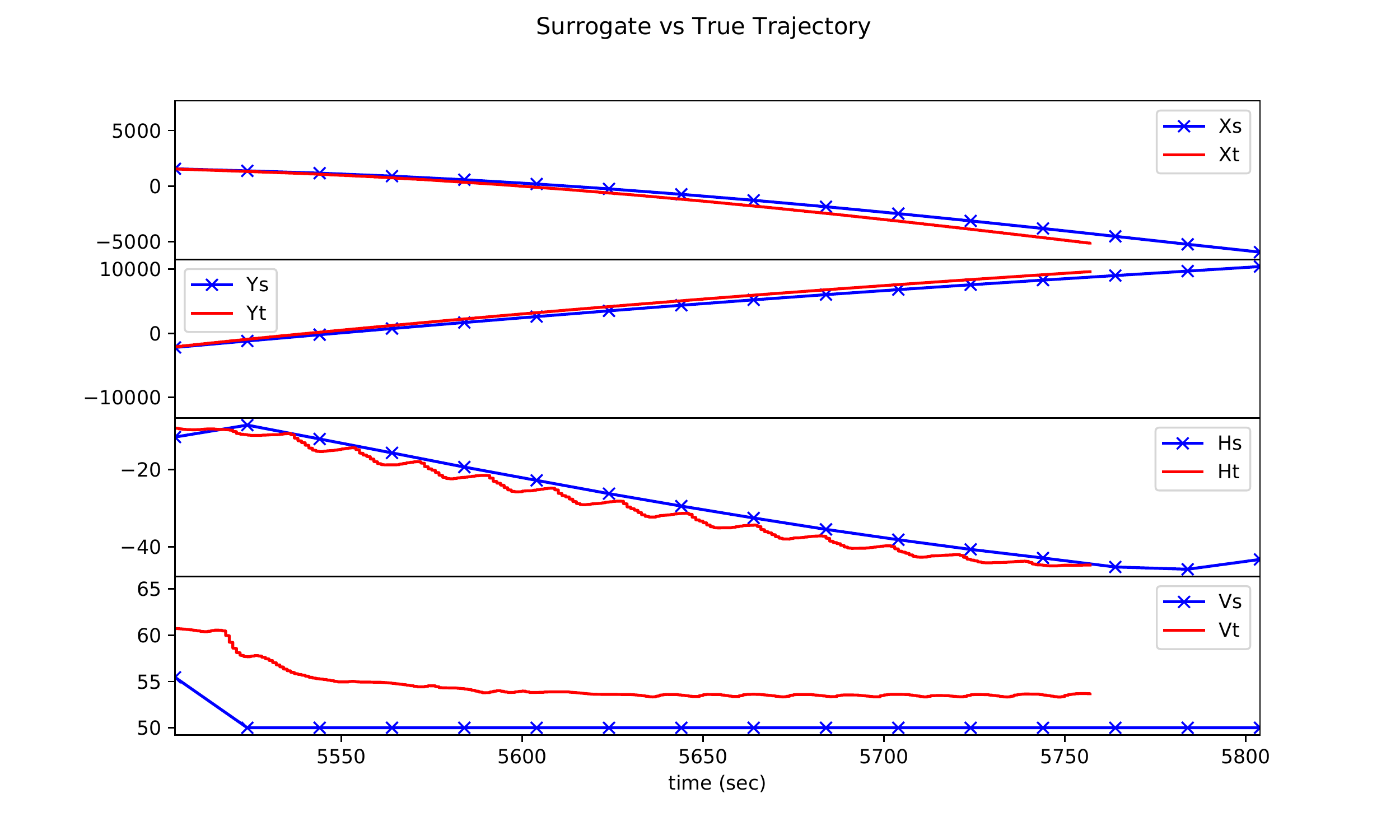}
        \caption{Accumulation of the error in the trajectory.}
        \label{fig:error_2}
    \end{subfigure}
    \caption{Transition error due to mismatch between the surrogate and the ground truth transitions.}
\end{figure}

This section evaluates how the demonstrator can closely follow the commanded trajectory. 
We set four experiments designed to force the Agent to demonstrate different behaviors.
These experiments include the following scenarios: Intruder crossing the planned route from left to right, from right to left, Intruder flying a head-on course, and Intruder flying the same course.

In all the scenarios, the vehicle's controller compensated for the discrepancy in the physical capabilities and surrogate assumptions.
Figures \ref{fig:traj_cross} and \ref{fig:traj_headon} demonstrate that the vehicle was following the waypoints compensating for the distance error and the time of arrival error.
\section{Conclusion}
\label{sec:concl}

This work presented an automated conflict resolution system trained as a Reinforcement Learning framework. The avoidance policy was trained offline on a surrogate task and stored as a neural network model.
The run-time RLCAS system carries the pre-computed solution library and provides a quick and efficient avoidance route to mitigate a potential traffic conflict.
All optimization is done offline and off the board of the flying vehicle, which allows running it on a low-power node.
This system would be an excellent fit for the small-scale fast-flying UAVs incapable of carrying a run-time path-planning solver.

The system is integrated with ROS, which allows transitioning from the simulations to the physical demonstrator. The live demonstration at Grant County International Airport, located six miles northwest of Moses Lake in Grant County, Washington, in 2021 generated flight data that will be analyzed and published in a separate paper.
With a relatively high number of failures, the RLCAS requires an assurance monitor running on top. We outlined two prospective assurance methods for the RLCAS system -- internal surrogate validation and external run-time monitoring. You may find additional information about the assurance in a separate paper \cite{cofer2022flight}.

Most importantly, this work provides an alternative methodology for solving a dynamic trajectory generation task. Diversification of the critical components is an essential concept for risk mitigation for the entire system.

\section*{ACKNOWLEDGMENT}
The authors thank Alex Chen and Michael McGivern from Boeing for helping with data collection efforts.

\bibliographystyle{unsrt}
\bibliography{main}

\medskip

\small

\appendix
\section{Appendix}
\label{sec:appendix}

\begin{figure}[h]
  \centering
  \hspace*{-10pt}
  \includegraphics[trim=50 0 0 0,clip, width=1.1\linewidth]{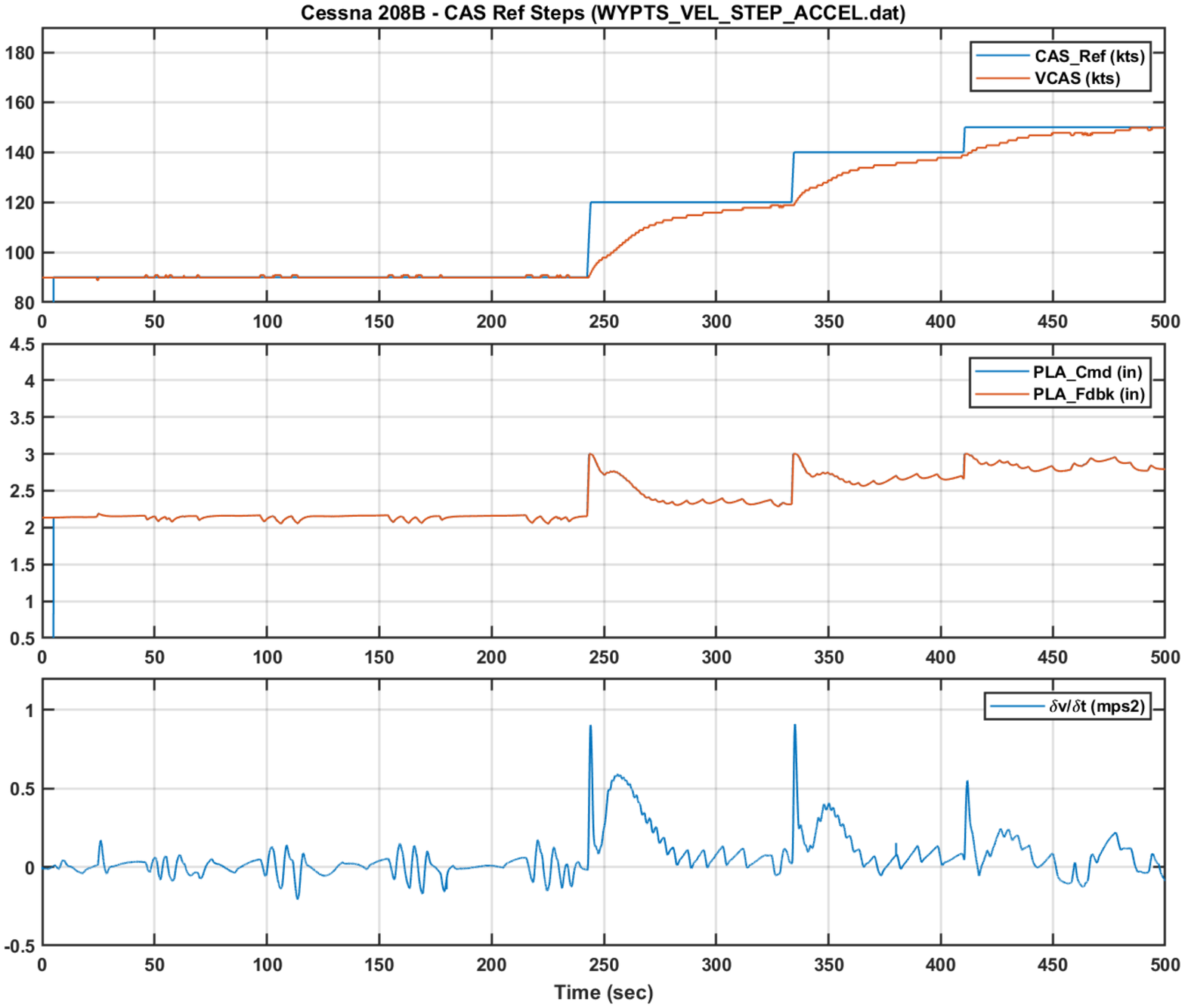}
    \caption{Airspeed and longitudinal acceleration response of the demonstrator to commanded step-inputs. }
    \label{fig:accel_data}
\end{figure}

\begin{figure}[h]
  \centering
  \hspace*{-10pt}
  \includegraphics[trim=50 0 0 0,clip, width=1.1\linewidth]{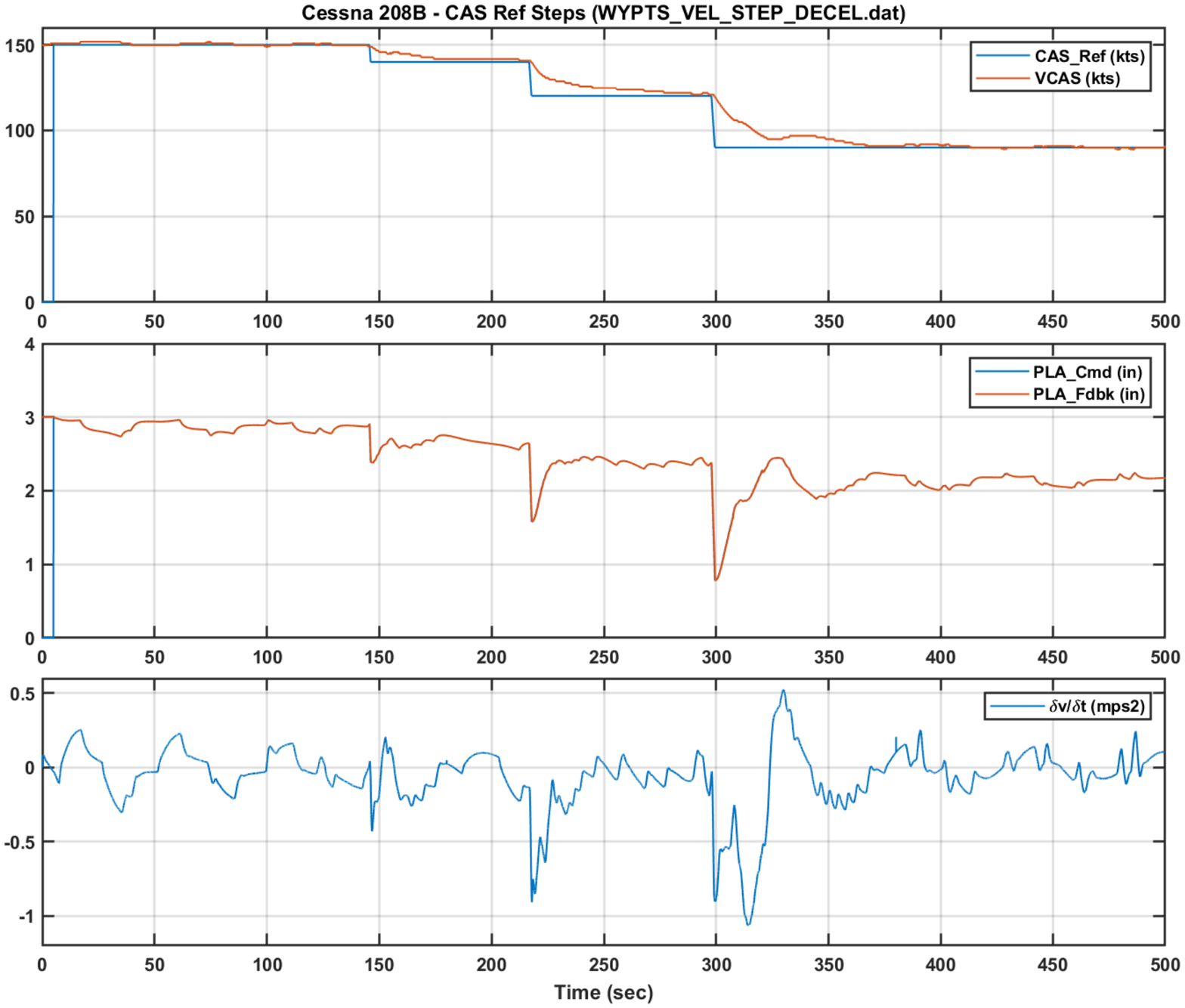}
    \caption{Airspeed and longitudinal deceleration response of the demonstrator to commanded step-inputs.}
    \label{fig:decel_data}
\end{figure}

\begin{table}[h]
    \centering
    \begin{tabular}{||c | c | c||} 
    \hline
    Parameter & Range & Units \\
    \hline
    Time step $dT$ & 1.0  & sec \\
    Max time $t_\text{MAX}$ & 300.0  & sec \\
    Distance range & -10000 .. 10000  & m \\
    Horizontal separation & 1000  & m \\
    \hline
    \end{tabular}
    \caption{Surrogate environment simulation parameters.}
    \label{tbl:sim_param}
\end{table}

\begin{table}
    \centering
    \begin{tabular}{||c | c ||} 
    \hline
    Condition & Reward \\
    \hline
    Distance violation & -100 \\
    No violation, missing the original path & -10 \\
    No violation, returning to original path & 100 \\
    \hline
    \end{tabular}
    \caption{Surrogate environment reward function.}
    \label{tbl:sim_reward}
\end{table}

\begin{algorithm}
\SetAlgoLined
Reset to random initials\;
 \While{$t<t_{max}$}{
    Update Intruder vehicle:
    \begin{itemize}
        \item Call vehicle controller to compute the control
        \item Set control signal to the vehicle
        \item Call vehicle update to move the vehicle
    \end{itemize}
    Update agent vehicle:
    \begin{itemize}
        \item Pass control signal to the vehicle
        \item Call vehicle update to move the vehicle
    \end{itemize}
    Evaluate the state: 
    \begin{itemize}
        \item Conflict
        \item Termination
        \item Score
    \end{itemize}
  \KwRet{\text{Observation, Reward, Termination}}\
 }
 \caption{Surrogate simulation update}
 \label{alg:updates}
\end{algorithm}

\end{document}